\documentclass[conference]{IEEEtran}

\IEEEoverridecommandlockouts

\usepackage{multirow}

\AtBeginDocument{%
  \providecommand\BibTeX{{%
    \normalfont B\kern-0.5em{\scshape i\kern-0.25em b}\kern-0.8em\TeX}}}

\usepackage{cite}
\usepackage{amsmath,amssymb,amsfonts}
\usepackage{algorithmicx}
\usepackage{graphicx}
\usepackage{textcomp}
\usepackage{xcolor}
\usepackage[export]{adjustbox}
\usepackage[outdir=./]{epstopdf}
\usepackage{amsmath,amsfonts}
\usepackage{amssymb}
\usepackage{graphicx}
\usepackage{enumerate}
\usepackage{textcomp}
\usepackage{xcolor}
\usepackage{url}
\usepackage{balance}
\usepackage{mathrsfs}

\usepackage{graphicx}
\usepackage{textcomp}
\usepackage{xcolor}

\usepackage{amsthm}
\usepackage{floatrow}

\theoremstyle{definition}

\usepackage{algorithm} 
\usepackage{algpseudocode}

\usepackage[tight,footnotesize]{subfigure}
\usepackage{caption}
\usepackage{color,soul,xspace}

\DeclareMathAlphabet\mathbfcal{OMS}{cmsy}{b}{n}

\newcommand{\mourmeth}{\text{MDTD}}
\newcommand{\ourmeth}{$\mourmeth$\xspace}

\def\BibTeX{{\rm B\kern-.05em{\sc i\kern-.025em b}\kern-.08em
    T\kern-.1667em\lower.7ex\hbox{E}\kern-.125emX}}

\title{Multi-Dictionary Tensor Decomposition}

\begin{document}

\author{\IEEEauthorblockN{Maxwell McNeil and Petko Bogdanov}
\IEEEauthorblockA{\textit{Computer Science, University at Albany- SUNY} \\ 
\{mmcneil2,pbogdanov\}@albany.edu}
}
\maketitle

\begin{abstract}
Tensor decomposition methods are popular tools for analysis of multi-way datasets from social media, healthcare, spatio-temporal domains, and others. Widely adopted models such as Tucker and canonical polyadic decomposition (CPD) follow a data-driven philosophy: they decompose a tensor into factors that approximate the observed data well. In some cases side information is available about the tensor modes. For example, in a temporal user-item purchases tensor a user influence graph, an item similarity graph, and knowledge about seasonality or trends in the temporal mode may be available. Such side information may enable more succinct and interpretable tensor decomposition models and improved quality in downstream tasks.

We propose a framework for Multi-Dictionary Tensor Decomposition (MDTD) which takes advantage of prior structural information about tensor modes in the form of coding dictionaries to obtain sparsely encoded tensor factors. 
We derive a general optimization algorithm for MDTD that handles both complete input and input with missing values. Our framework handles large sparse tensors typical to many real-world application domains. We demonstrate MDTD's utility via experiments with both synthetic and real-world datasets. It learns more concise models than dictionary-free counterparts and improves (i) reconstruction quality ($60\%$ fewer non-zero coefficients coupled with smaller error); (ii) missing values imputation quality (two-fold MSE reduction with up to orders of magnitude time savings) and (iii) the estimation of the tensor rank. MDTD's quality improvements do not come with a running time premium: it can decompose $19GB$ datasets in less than a minute. It can also impute missing values in sparse billion-entry tensors more accurately and scalably than state-of-the-art competitors.
\end{abstract}


\section{Introduction}

Tensors are multi-way arrays that generalize matrix data to higher number of ``dimensions''~\cite{tensor_tutorial}. The ability of tensors to accurately model the complex relationships present in many datasets has rendered them applicable in signal processing~\cite{sidiropoulos2000blind}, machine learning~\cite{papalexakis2012parcube},  chemometrics~\cite{PARAFAC_CPD}, and other fields.
Similar to matrices, low rank decomposition models for tensors are common ways of finding patterns in multi-way data. 
Popular approaches like the Canonical polyadic decomposition (CPD)~\cite{PARAFAC_CPD} and Tucker decomposition~\cite{tucker1966some} learn directly from data without additional modeling assumptions. 
In many settings prior knowledge about the data generation process may also be available, for example, seasonality in a temporal mode or a network associating individuals in a user mode. In addition, downstream applications such as data imputation, clustering and anomaly detection may benefit from imposing structure in the decomposition. Such considerations have given rise to modifications to the original CPD and Tucker models that have produced state-of-the-art performance in missing values imputation within a Bayesian framework~\cite{BGCP,BATF}, improved community detection for on/off~\cite{gorovits2018larc}, periodic~\cite{zhang2019PERCeIDs} or bursty self-exciting behavior~\cite{MYRON}, and other tasks.


\begin{figure} [t]
    \centering
     \includegraphics[width=1\textwidth]{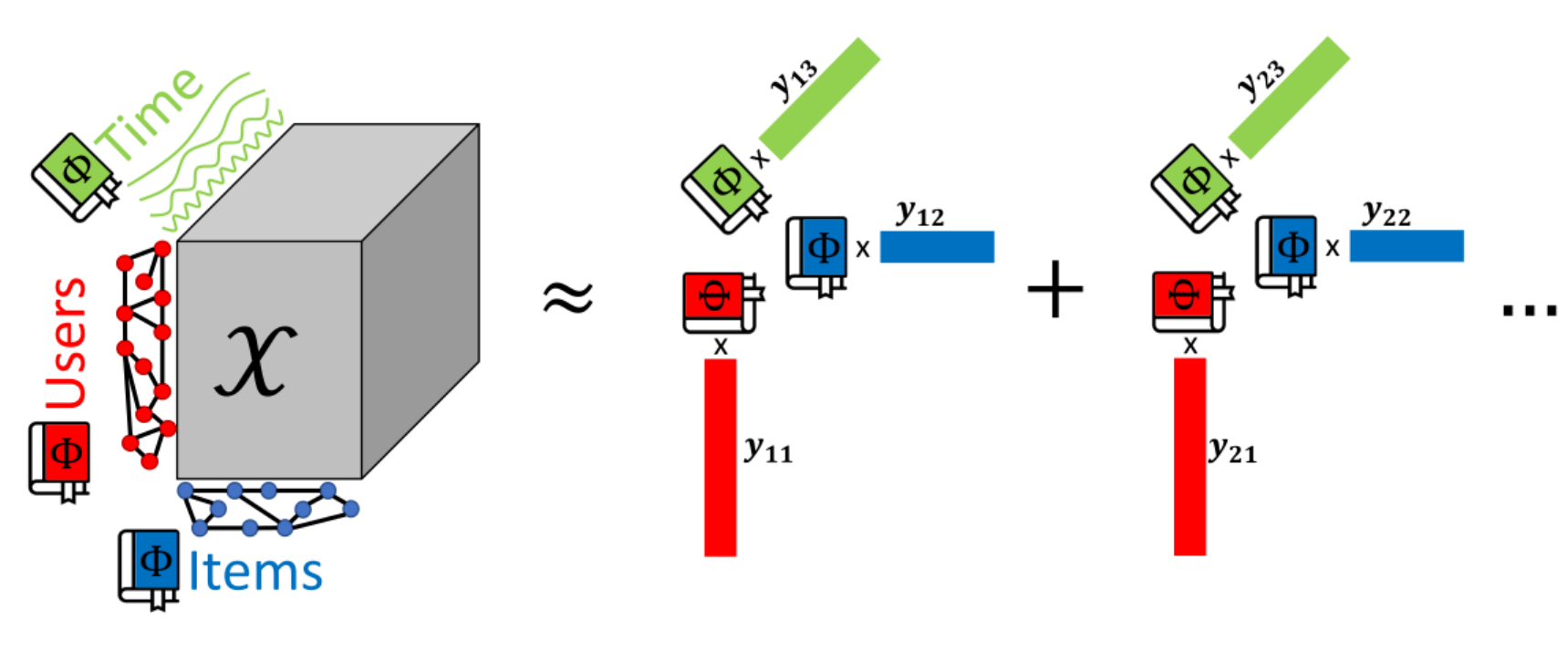}
     \caption{\footnotesize The key idea behind the dictionary-based tensor decomposition model (MDTD) through a user-item-time example. MDTD can utilize graph-based dictionaries $\Phi$ for the user and item modes (e.g. Graph Fourier Transform or Graph wavelets) and a temporal dictionary such as a Fourier dictionary for the temporal mode. The decomposition is similar to CPD decomposition in that it is a sum of rank-one factor tensors, with the key difference that factors are represented as encoding $y_{ij}$ through the corresponding dictionary. }
     \label{fig:overview}  
\end{figure}

Most methods employ regularization to build prior knowledge into the factorization model imposing different forms of structure: sparsity, periodicity and others. An alternative approach is to employ sparse coding for tensor factors via dictionaries~\cite{dic_tensor_decomp}. Such sparse coding techniques utilize fixed dictionaries and have been widely adopted in signal and graph signal processing~\cite{tennetiTSP2015,ortega2018graph}, computer vision~\cite{fadili2009inpainting}, machine learning~\cite{gregor2010learning} and data analytics~\cite{TGSD}. The ubiquitous applications of such methods have also given rise to some standard analytical dictionaries for time series (Fourier, Ramanujan, splines)~\cite{tennetiTSP2015}, graphs (graph Fourier and graph wavelets)~\cite{ortega2018graph}, and images (wavelets, ridgelets, curvelets)~\cite{fadili2009inpainting}. Employing such dictionaries for tensor data promises to enable succinct, interpretable and efficient-to-learn models. 

We introduce a multi-dictionary tensor factorization (MDTD) framework that employs fixed dictionaries for joint sparse coding of the tensor factors. The key idea of our model is illustrated via a user-item-time example tensor in Fig.~\ref{fig:overview}. Given prior knowledge in the form of user and item graphs as well as expectation about periodic behavior in time, we propose to employ corresponding dictionaries $\Phi$ to sparsely encode factors in a CPD-like model. For the example in the figure, we can employ a Graph Fourier Transform (GFT) dictionary for the modes with graph side information and a periodic dictionary for the temporal mode. 
The model is applicable to higher order tensors with any subset of modes endowed with side information, as well as to other kinds of side information and corresponding dictionaries. 
We propose a general optimization solution for MDTD and evaluate it on multiple tensor datasets. We demonstrate that when the side information captured by the dictionaries is well aligned with the data in the tensor, our approach enables i) orders of magnitude reduction in the model size compared to CPD and Tucker, while running in comparable time and ii) enables improved quality in several downstream tasks.

Our contributions in this paper are as follows:

\noindent{\bf $\bullet$ Generality and Novelty:} We propose a general tensor decomposition framework \ourmeth, which to the best of our knowledge, is the first to decompose a tensor via sparse multi-dictionary coding. 

\noindent{\bf $\bullet$ Parsimony and Scalability:} 
\ourmeth produces interpretable and concise representations of both real-world and synthetic tensors scaling similar to simple decomposition models and better than more complex ones. \ourmeth processes a $19GB$ tensors in $1$ min and can impute missing values in tensors with billions more entries than what competitors can handle.

\noindent{\bf $\bullet$ Applicability and Accuracy:} 
We demonstrate \ourmeth's utility for succinct tensor representation, rank estimation and missing value imputation. Its quality dominates baselines across applications and datasets. In some cases \ourmeth achieves higher accuracy and a 100x speed-up compared to the fastest baseline.


\section{Related Work}



{\noindent \bf Sparse dictionary coding} models data as a sparse combination of dictionary bases (atoms). It is widely used in signal processing~\cite{zhang2015survey}, graph signal processing~\cite{Shuman_2013}, time-series analysis~\cite{tennetiTSP2015}, computer vision~\cite{wright2008robust}, and others.   Recent work has utilized Kronecker products of multiple graph dictionaries to allow for filtering to applied to vectorizations of tensors data~\cite{stanley2020multiway}, but do not offer a way to directly decompose and encode higher-order data. 
Our work generalizes dictionary coding to multi-mode tensors, hence the relevant literature on sparse dictionary coding is complementary to our approach.

\noindent{\bf Tensor decomposition} is a well-studied topic with multiple competing methods among which CPD~\cite{PARAFAC_CPD} and Tucker\cite{tucker1966some} stand out as the most fundamental models.  
There also exist many approaches for missing tensor value imputation tailored to specific applications, e.g., road traffic~\cite{BATF,BGCP} and images~\cite{CP-WOPT,TRLRF}. We compare to such methods in our experimental evaluation. Some methods perform coupled tensor factorization \cite{tensor_tutorial} by enforcing sharing of factors with an additional coupled tensors (or matrices). The coupled data can be viewed as side information, however the goal is to (co-)factorize them, while in our setting dictionaries are used as encoding basis rendering the two problem settings unrelated. 

A multitude of extensions to basic tensor decomposition constrain the models for desirable properties. Some perform dictionary learning within the factors via alternating optimization where either the dictionary or the encoding is fixed~\cite{visual_tensor_complete,tucker_dic_learning}. 
For example, the authors of~\cite{tucker_dic_learning} constrain a Tucker decomposition to have a sparse core tensor in order to learn dictionaries. The authors of \cite{visual_tensor_complete} learn a dictionary learning from image data to facilitate the imputation of missing pixel information.
Dictionary learning is complementary to our work as we can leverage learned dictionaries within our model for the specific datasets they are designed for. We focus our experiments on analytical dictionaries generated from a mathematical model (e.g. discrete Fourier transform) which typically generalize better than their learned counterparts~\cite{rubinstein2010dictionaries}.

Dictionaries are also employed to regularize factors in CPD decomposition to fit expected properties~\cite{MYRON,zhang2019PERCeIDs,gorovits2018larc}. For example the authors of~\cite{MYRON} and~\cite{zhang2019PERCeIDs} regularize temporal mode factors to exhibit bursty and periodic behavior respectively for community detection. Our approach differs from the above works in that we employ dictionaries to directly encode factors rather than to regularize them, we jointly utilize multiple dictionaries across multiple modes, and our method is a general factorization model rather than a community detection method. 
The authors of~\cite{dic_tensor_decomp} model CPD factors of image data via a restricted one-atom-per-factor encoding. Our method can be considered a sparse coding generalization which jointly employs multiple dictionaries.




\noindent{\bf Multi-dictionary approaches} decompose input data by employing a combination of dictionaries. The authors of~\cite{multi_dic_sen} propose a matrix factorization approach for spectral unmixing of images based on integrating multiple dictionaries into one super dictionary. This approach is complementary to ours as we can leverage such composite dictionaries when the application necessitates it. A more closely-related work~\cite{TGSD} decomposes a temporal graph signal matrix by employing a temporal dictionary and a graph dictionary. Our method can be viewed as a generalization to multi-way datasets. 
We experimentally demonstrate the advantages of our method over this baseline employed on tensor slices.

%
%
\section{Preliminaries}
Before we define our problem of dictionary-based tensor decomposition (\ourmeth), we first introduce necessary preliminaries and notation. The input to our problem is a tensor $\mathcal{X}$, which is a multi-dimensional array of real numbers. We present the problem and our solutions in the context of three-way tensors for simplicity, however, both generalize seamlessly to higher order tensors. We will work with tensors of the following shape $\mathcal{X}\in\mathbb{R}^{(I \times J \times T)}$, where $I$, $J$ and $T$ are the dimensions of the modes. 

\noindent{\bf CPD decomposition.} \ourmeth can be viewed as a dictionary-based extension of the CPD decomposition of the form:  
\begin{equation}
\begin{aligned}
\footnotesize
     \mathcal{X} = \sum_{i=1}^k \mathcal{H}_i =  \sum_{i=1}^k a_i \boxtimes b_i \boxtimes c_i, 
    \end{aligned}
\end{equation}
where $\boxtimes$ denotes the tensor outer product and $\mathcal{H}_i$ are rank-one tensors obtained from outer tensor products of individual factors $a_i\in\mathbb{R}^{I},b_i\in\mathbb{R}^{ J },$ and $c_i\in\mathbb{R}^{T}$. If we stack $k$ factor vectors $a_i,b_i,$ and $c_i$ into matrices $A\in\mathbb{R}^{I \times k},B\in\mathbb{R}^{ J \times k },$ and $C\in\mathbb{R}^{T \times k}$ respectively, we can express this relationship concisely as: $\mathcal{X} = [[A,B,C]]$.
An in-depth introduction of CPD and other tensor models is available in~\cite{tensor_tutorial}.

\noindent{\bf Sparse dictionary coding} or sparse representation modeling~\cite{rubinstein2010dictionaries} assumes that the data can be represented via a linear combination of a few atoms from an appropriately-chosen pre-specified dictionary $\Phi$, where both analytical and dictionaries learned from data can be employed. In its general form sparse coding solves the following problem: 

$$\min_y f(y)~~\text{s.t.}~~ x=\Phi y,$$
where $x$ is an input signal, $y$ is its encoding and $f(y)$ is a sparsity promoting function often instantiated as an $L_1$ norm.  
\section{Problem Formulation and Solution}
In many real-world application there is a structural information associated with tensor modes. Consider, for example, users (mode 1) watching streams (mode 2) over time (mode 3) on a stream service such as Twitch. Such data can be represented by a binary tensor $\mathcal{X}\in\mathbb{R}^{(I \times J \times T)}$. It is easy to imagine that users may be associated within a friendship network and streams within a topical similarity network. More over the communities within those networks (friendship groups interested in streams featuring similar games) will likely stream based on regular daily/weakly patterns. \emph{How can we leverage this rich structural information to learn a succinct, interpretable, and meaningful representation of the data?} 

We propose to represent a tensor with structural side information through a CPD-like dictionary-based decomposition: 

\begin{equation*}
\begin{aligned}
\footnotesize
    \mathcal{X} =\sum_{n=1}^k \Phi_1 y_{n1}  \boxtimes \Phi_2 y_{n2} \boxtimes \Phi_3 y_{n3} = [[\Phi_1Y_1,\Phi_2Y_2,\Phi_3 Y_3]], 
    \end{aligned}
    \label{equation:basic_dic_tensor}
\end{equation*}
where prior knowledge in each mode is incorporated as a model-specific dictionary $\Phi_i$ and the sparse encoding of the input data through dictionaries is in matrices $Y_i$. Fitting the input data to such a model results in the following problem:  


\begin{equation*}
\begin{aligned}
\footnotesize
    \min_{Y_1,Y_2,Y_3} \frac{1}{2}||\mathcal{X} - [[\Phi_1Y_1,\Phi_2Y_2,\Phi_3 Y_3]] ||_F^2 + \sum_{i=1}^3 \lambda_i\left\|Y_i
    \right\|_1,
    \end{aligned}
\end{equation*}
where the first term is the data fit and the second term encourages sparsity in encodings $Y_i$ in the form of an $L_1$ regularization. This form of sparsity is typical when using dictionaries to avoid overfitting and ill-posed problems. Increasing the sparsity balance parameters $\lambda_i$ encourages sparser solutions for corresponding modes and allows us to control the complexity/size of the learned model. 

In many applications tensor data is sparsely populated and features missing/unobserved values. To allow our model to handle such scenarios we also introduce a zero-one mask $\Omega$ which is a tensor of the same size as $\mathcal{X}$, to prevent the model from fitting missing values. Our overall \ourmeth objective is:   

\begin{equation}
\begin{aligned}
\footnotesize
    \min_{Y_1,Y_2,Y_3} \frac{1}{2}||\Omega\boxdot(\mathcal{X} - [[\Phi_1Y_1,\Phi_2Y_2,\Phi_3 Y_3]]) ||_F^2  + \sum_{i=1}^3 \lambda_i\left\|Y_i
    \right\|_1,
    \end{aligned}
    \label{final_obj_01}
\end{equation}
where $\boxdot$ denotes the element-wise product. It is important to note that if a dictionary (or side information) is not available for some of the modes in a given application, a trivial identity dictionary $\Phi_i=I$ and a corresponding 0 sparsity  cost ($\lambda_i=0$) will allow that mode to be fit as in a regular CPD model.

\subsection{Optimization}

Optimizing the objective from Eq.~\ref{final_obj_01} directly with respect to all factors $Y_i$ is not trivial. We employ an iterative one-factor-at-a-time approach similar to solvers for CPD. Specifically, we employ an ADMM approach to partition the problem into sub-problems with closed-form solutions.
We first introduce intermediate variables $Z_i = Y_i$ and $\mathcal{X}=\mathcal{D}$, resulting in: 

\begin{equation}
\begin{aligned}
\footnotesize
    \min_{Y_i,Z_i,\mathcal{D}} \frac{1}{2}||\mathcal{D} - [[\Phi_1Y_1,\Phi_2Y_2,\Phi_3 Y_3]] ||_F^2   + \sum_{i=1}^3 \lambda_i\left\|Y_i
    \right\|_1 \\
    + \lambda_d\left\| \Omega\boxdot(\mathcal{D}- \mathcal{X}) \right\|_F^2 
    ~s.t.~
     Y_1=Z_1,Y_2=Z_2,Y_3=Z_3,\mathcal{X}=\mathcal{D}
    \end{aligned}
\end{equation}

With some algebraic transformations, the corresponding Lagrangian form of the objective is:
\begin{equation}
\begin{aligned}
\footnotesize
    \min_{Y_i,Z_i,\mathcal{D},\Gamma_i} \frac{1}{2}||\mathcal{D} - [[\Phi_1Y_1,\Phi_2Y_2,\Phi_3 Y_3]] ||_F^2 +  \sum_{i=1}^3 \lambda_i\left\|Z_i
    \right\|_1 \\
    + \sum_{i=1}^3\frac{\rho_i}{2}\left\|Y_i- Z_i
    + \frac{\Gamma_i^\tau}{\rho_i}  \right\|_F^2  + \lambda_d\left\| \Omega\boxdot(\mathcal{D}- \mathcal{X}) \right\|_F^2.
    \end{aligned}
\end{equation}

Our ADMM optimization updates one variable while keeping the rest fixed. This ensures simple and tractable updates. Just like the problem definition, we derive the solutions for tensors of $3$ modes for simplicity, however, the algorithm generalizes to higher order tensors. We assume that side information in the form of dictionaries is available for each of the tensor modes, however, modes lacking side information can be thought of as employing the canonical basis dictionary, i.e. $\Phi=I$, and the optimization for those modes can be performed similar to those in standard CPD.

To simplify the derivation we will employ the following shorthand matrices: $A=\Phi_{j} Y_{j},B=\Phi_l Y_l$ where $j<l$. A key step in learning our decomposition involves an unfolding on the tensor along a specific mode to produce a matrix. To achieve this, slices of the tensor are vectorized and stacked along all modes except one. For example, the unfolding of tensor $\mathcal{X}$ on its first mode is defined and denoted as: $X_1=[vec(\mathcal{X}(1,:,:)),vec(\mathcal{X}(2,:,:)),...,vec(\mathcal{X}(I,:,:))]$, where $X_1\in \mathbb{R}^{(JT \times I )}.$ We will employ unfoldings for both the input $\mathcal{X}$ and approximation $\mathcal{D}$ tensors, where mode $i$ unfoldings are denoted $X_i$ and $D_i$ respectively. With these definitions we are ready to derive the individual updates.

\noindent\textbf{Updates for $Y_i$'s.} The subproblem with respect to the encoding matrices $Y_i$ can be written as:    

\begin{equation}
\begin{aligned}
\footnotesize
    \min_{Y_i} \frac{1}{2}||D_i^T - \Phi_i Y_i (B \odot A)^T ||_F^2 +\frac{\rho_i}{2}\left\|Y_i - Z_i + \frac{\Gamma_i^\tau}{\rho_i} \right\|_F^2,
    \end{aligned}
\end{equation}
where $\odot$ is the Khatri-Rao product~\cite{tensor_tutorial}. The derivative with respect to $Y_i$ is as follows:

\begin{equation}
\begin{aligned}
\footnotesize
    \Phi_i^T \Phi_i Y_i [B^T B\boxdot A^TA]- \Phi_i^T D_i^T(B \odot A)  + \rho_iY_i - \rho_iZ_i + \Gamma_i^\tau,
    \end{aligned}
\end{equation}
where $\boxdot$ denotes element-wise product and to obtain this form we have used the fact that $(B \odot A)^T(B \odot A)=B^T B\boxdot A^TA$, whose proof can be found in~\cite{tensor_tutorial} section 7 part A.

After setting the gradient to zero and rearranging we obtain:

\begin{equation}
\begin{aligned}
\footnotesize
\Phi_i^T \Phi_i Y_i [B^T B\boxdot A^TA] +\rho_i Y_i = \Phi_i^T D_i^T(B \odot A) + \rho_iZ_i - \Gamma_i^{\tau}\\
    \end{aligned}
\end{equation}

To solve for $Y_i$ we compute the eigenvalue decomposition of $\Phi_i^T\Phi_i$ and $[B^T B\boxdot A^TA]$. It is important to note that since we are typically interested in low-rank decomposition in practical applications, the decomposition of $[B^T B\boxdot A^TA] \in \mathbb{R}^{k \times k}$ is fast in practice.
For $\Phi_i^T\Phi_i$ we can also obtain a fast solution as we do not need to compute its eigenvalue decomposition directly. 
Let $\Phi_i=E_d \Lambda_d^\frac{1}{2} V_d^T$ be the SVD decomposition of $\Phi_i$. Then, 
$$\Phi_i^T\Phi_i = (V_d \Lambda_d^\frac{1}{2} E_d^T)^T V_d \Lambda_d^\frac{1}{2} E_d^T =E_d \Lambda_d E_d^T.$$
Let also $[B^T B\boxdot A^TA]=E_{v} \Lambda_v E_{v}^T$ be represented by its eigendecomposition. Using the two definitions we obtain the following: 

\begin{equation}
\begin{aligned}
\footnotesize
E_d \Lambda_d E_d^T Y_i E_{v} \Lambda_v E_{v}^T +\rho_i Y_i  =\underbrace{\Phi_i^T D_i^T(B \odot A) + \rho_iZ_i - \Gamma_i^{\tau}}_{C}.
    \end{aligned}
    \label{eq:yi-evd}
\end{equation}
Multiplying both sides of Eq.~\ref{eq:yi-evd} by $E_d^T$ and $E_{v}$ on the left and right respectively, we obtain:
\begin{equation}
\begin{aligned}
\footnotesize
 \Lambda_d E_d^T Y_i E_{v} \Lambda_v+ \rho_i E_d^T Y_i  E_{v} = E_d^T C E_{v}
    \end{aligned}
\end{equation}

 Let $\textbf{p}_{d}$, $\textbf{p}_{v}$ be the vectors corresponding to the diagonal elements of $\Lambda_d$ and $\Lambda_v$ respectively. We can then derive the following closed-form solution for $Y_i$:

\begin{equation}
\begin{aligned}
\footnotesize
  Y_i =  E_{d} [(E_{d}^T C E_{v}) \oslash (\textbf{p}_{d}*\textbf{p}_{v}^T +\rho_i)]E_{v}^T,
    \end{aligned}\label{not_ortho_update}
\end{equation} 
where $\oslash$ denotes element-wise division.

Importantly, when working with an orthogonal dictionary $\Phi_i$ (e.g., DFT or GFT), we can obtain a simplified (non-SVD) update as follows:   

\begin{equation}
\begin{aligned}
\footnotesize
    Y_i = ( \Phi_i^T D_i^T(B \odot A)+ \rho_iZ_i - \Gamma_i^{\tau})(B^T B\boxdot A^TA +\rho_i I)^{-1}
    \end{aligned}
\end{equation}


{\noindent \bf Update for $Z_i$:}
The problem w.r.t. the proxy variables for the encodings $Z_i$ is as follows:
\begin{equation}
\begin{aligned}
   \underset{Z_i}{\mathrm{argmin}} \hspace{0.1cm} \lambda_i\left \| Z_i \right \|_1 + \frac{\rho_i}{2} \left \| Y_i-Z_i + \frac{\Gamma_i^\tau}{\rho_i} \right \|_F^2 \\ 
\end{aligned}
\end{equation}
Closed-form solution for this problem is available due to~\cite{Lin2013TheAL}:
\begin{equation}
\begin{aligned}
  Z_{i,jl} = sign\left (  H^{(i)}_{jl}\right )\times max\left ( \left | H^{(i)}_{jl}  \right | -\frac{\lambda_i}{\rho_i},0 \right ),  
\end{aligned}
\end{equation}
where $H^{(i)}= Y_i - \frac{\Gamma_i^\tau}{\rho_i}.$

{\noindent \bf Update for $\Gamma_i$:} We update the Lagrangian multipliers as follows:
\begin{equation}
\begin{aligned}
\Gamma_i^{\tau+1} =\Gamma_i^{\tau} +\rho_i\left ( Z_i-Y_i  \right ),
\end{aligned}
\end{equation}
where $\tau$ is equal to the number of the current iteration.

{\noindent \bf Update $\mathcal{D}$:} The sub-problem with respect to the full data (no missing values) reconstruction $\mathcal{D}$ is as follows: 

\begin{equation}
\begin{aligned}
\frac{1}{2}||\mathcal{D} - [[\Phi_1Y_1,\Phi_2Y_2,\Phi_3 Y_3]] ||_F^2 +\lambda_d\left\| \Omega \boxdot (\mathcal{D}- \mathcal{X}) \right\|_F^2
\end{aligned} \label{eq:dense_impute_form}
\end{equation}

By setting its gradient to zero we obtain the following update:   

\begin{equation}
\begin{aligned}
 \mathcal{D} = ([[\Phi_1Y_1,\Phi_2Y_2,\Phi_3Y_3]]+\lambda_d\Omega \boxdot  \mathcal{X})\oslash ( \mathcal{I}+\lambda_d\Omega).
\end{aligned} \label{eq:dense_impute_sol}
\end{equation}


\begin{center}
	\begin{algorithm}[!t]
		\footnotesize
		\caption{\ourmeth (with missing values)}
		\label{alg:opt}
		\begin{algorithmic}[1]
			\Require{ Input $\mathcal{X}$, mask $\Omega$, dictionaries $\Phi_i$, $k$, $\lambda_i$, $\rho_i$
			}
 \State  Initialize $Y_i=Z_i$ uniformly random,  and $\Gamma_i=0$ for all modes, set $\mathcal{D}=\mathcal{X}$


                \For{ \textit{i= 1 to \#modes}}
                \If{$\Phi_i^T \Phi_i \neq I$}
                
                  \State    $ E_{d,i} \Lambda_{d,i} E_{d,i}^T =\Phi_i^T\Phi_i$
                \EndIf
                \EndFor
    
                \While{ not converged }
                
                    \For{ \textit{i= 1 to \#modes}}

                    \State set $j\neq l\neq i$ and $j<l$
                    \State $A=\Phi_{j} Y_{j}$
                    \State $B=\Phi_l Y_l$
                    
                    \If{$\Phi_i^T \Phi_i=I$}

                    \State $Y_i = ( \Phi_i^T D_i^T(B \odot A)+ \rho_iZ_i - \Gamma_i^\tau )(B^T B\boxdot A^TA +\rho_i I)^{-1}$ 
                   
                   \Else

                   \State $  E_{v} p_v E_{v}^T= B^T B\boxdot A^TA $
                   
                   \State $C=\Phi_i^T D_i^T(B \odot A) + \rho_iZ_i - \Gamma_i^{\tau}$
                   
                    \State  $Y_i =  E_{d,i} [(E_{d,i}^T C E_{v}) \oslash (\textbf{p}_{d,i}*\textbf{p}_{v}^T +\rho_i)]E_{v}^T$
                    
                    \EndIf
                    
                    \State  $S_{f} = max(Y_{i,f})$,  for $f$ from $1$ to $k$

                     \State  $Y_i = Y_i \oslash S$

                    \State $H^{(i)}= Y_i - \frac{\Gamma_i^\tau}{\rho_i}.$
                    
                    \State  $Z_{i,jl} = sign\left (  H^{(i)}_{jl}\right )\times max\left ( \left | H^{(i)}_{jl}  \right | -\frac{\lambda_i}{\rho_i},0 \right )$
                    
                    \State $\Gamma_i^{\tau+1} =\Gamma_i^{\tau} +\rho_i\left ( Z_i-Y_i  \right )$

                     \EndFor

                    \State  $ \mathcal{D} = ([[S \boxdot \Phi_1Y_1,\Phi_2Y_2,\Phi_3Y_3]]+\lambda_d\Omega \odot  \mathcal{X})\oslash ( \mathcal{I}+\lambda_d\Omega)$

                      \State $\tau \leftarrow \tau+1$ 
                \State Convergence condition: $\left | f^{t+1}-f^{t} \right | \leq \varepsilon $, where $f^{t+1}$ and $f^{t}$ are the objective values of Eq.~\ref{final_obj_01} at iterations $t+1$ and $t$.
                
                 \EndWhile      
		\end{algorithmic}     
	\end{algorithm}
\end{center} 
 
\subsection{ MDTD  algorithm and complexity } 


We present the overall optimization algorithm in the case of tensors with missing values in Alg.~\ref{alg:opt}. We first initialize all variables (Step 1) and pre-compute eigenvalue decompositions of $\Phi^T\Phi$ for non-orthogonal dictionaries (Steps 2-6). In the main loop of the algorithm (Steps 7-28) we iteratively update each mode's factors (Steps 8-24) and update the missing value imputation matrix (Step 25) until convergence. In Steps 9-11 we compute the factors for modes that are not currently being updated through their respected dictionaries $\Phi$ and coding matrices $Y$. The updates for the factor of a given $Y_i$ depend on whether the corresponding dictionary $\Phi_i$ is orthonormal. If $\Phi_i$ is orthonormal, we have a direct update (Step 13). The update for non-orthonormal dictionaries $\Phi_i$ employ the pre-computed eigendecompositions of their dictionaries and require three steps (15-17) based on our derivations in Eqs.~\ref{eq:yi-evd}-\ref{not_ortho_update}. 

We normalize learned factors in Steps $19-20$ by dividing each factor by its maximum value. Similar normalization is commonly used in CPD algorithms to ensure that the scale of each factor is bounded~\cite{normalzing_factors}. Finally, we update proxy variables and Lagrangian coefficients following the ADMM updates in Steps (19-23). When the input tensor does not have missing values, or their imputation is not necessary (i.e., we simply need a decomposition), we omit step 25 and simply replace all unfoldings $D_i$ with the unfolding of the input tensor $X_i$ elsewhere in the algoithm.  
The three steps of Alg.~\ref{alg:opt} which dominate the computational complexity are (i) the matrix inversion in step $13$, which runs in $O(k^3)$ (ii) the tensor reconstruction in step $25$ $[[S\boxdot \Phi_1Y_1,\Phi_2Y_2,\Phi_3Y_3]]$ involving the Khatri–Rao product of three matrices of sizes $I\times k$,$J\times k$ and $J\times k$ with complexity $O(IJTk)$
and (iii) the product $\Phi_i^T D_i^T(B \odot A)$ in steps $13$ and $16$. Let $\Phi_i^T$ be $p_i\times m_i$, $D_i^T$ be of size $m_i \times m_j m_l $ and $(B \odot A)$ be $ m_j m_l \times k$, then the complexity of the latter step is $O(p_i m_i m_j m_l+p_i m_j m_l k)$ if one performs $\Phi_i^T D_i^T$ first or $O(p_i m_i k+m_i m_j m_l k)$ if $ D_i^T(B \odot A)$ is performed first. 
The model rank $k$ and the number of dictionary atoms $p_i$ are the two hyperparameters that directly affect the overall complexity. The typical motivation behind tensor decomposition is that real-world tensors are often of low rank, i.e.,  ($k<p_i$). Assuming also that the number of atoms is of the same order as the size of the associate tensor mode ($p_i = O(m_i)$) leads to an asymptotic running time similar to dictionary-free updates such as ALS-based CPD. 
Reconstructing the full tensor $\mathcal{D}$ with missing values in Step 25, requires materializing a potentially dense large tensor even if the input and the number of missing values are relatively sparse. We discuss an alternative scalable solution for this step in for the case of large sparse tensors in the following section.



\begin{table*}[t]
\footnotesize
\setlength\tabcolsep{3 pt}
\centering

 \begin{tabular}{|c|c|c|c|c|c|c|c|c|c|c|c|c|c|c|c|c|c|c|c|c|c|c|c|c|c|} 
 \hline
    \multicolumn{6}{|c|}{\bf Dataset statistics} & \multicolumn{3}{|c|}{\ourmeth}&  \multicolumn{3}{|c|}{TGSD} &  \multicolumn{3}{|c|}{CPD} &  \multicolumn{3}{|c|}{Tucker} &  \multicolumn{3}{|c|}{TT}\\
 \hline
 { Dataset} & {\bf $m_1$}  & {\bf $m_2$} & Prior & {\bf $m_3$} & Prior  &  SSE & NNZ & time &  SSE & NNZ & time   &  SSE & NNZ & time &  SSE & NNZ & time &  SSE & NNZ & time   \\
    \hline
Syn &  200  & 300 & Graph & 400 & Period  
& \textbf{500} &  \textbf{1045} & 2   & 522 & 508K & 35
 & 605  & 8829  & 2.4
 & 681  & 8100  & \textbf{.5} & 607 & 80K & 1.5

\\
 
   \hline
RM &  94& 94 & Graph & 719 & Hours  
& \textbf{7M} &  \textbf{44K}   & .38 & 8M &  267K & 12
 & \textbf{7M}  & 64K  & .49
 & \textbf{7M}  & 100K  & \textbf{.31}  & \textbf{7M} & 180K & 1.2
\\

 \hline
Crime  & 77  & 24 & Hours & 6186 & Days  & \textbf{2.19K}  &  {\textbf{3794}} & .23  & 2.32K & 7k & 89 & 2.20K  & 6K  & \textbf{.04} & 2.20K & 6K  &  .69 & 3.08K & 14M & 3.6
\\
\hline
 
 Twitch-S & 5000 & 300 & Graph & 50-500 & Hours 
 & \textbf{1.7M} &  \textbf{9K} & 17* 
 & 16M & 8M & 9K
 & 1.8M  & 55K &  \textbf{1}*
 &  1.8M  & 56K  & 6*
 & 16M & 225M & 90
\\  \hline

 Twitch-M & 8000 & 500 & Graph & 50-500 & Hours 
 & \textbf{5M} &  \textbf{14K} & 36*
& \textbf{5M}  & 455K & 76K
 & \textbf{5M}  & 87K &  \textbf{3}*
 & \textbf{5M}  & 88K  & 9*
  & \textbf{5M} & 700K & 178*
\\  \hline
 
  Twitch-L & 8000 & 3000 & Graph & 500 & Hours 
 & \textbf{20M} &  \textbf{10K} & 71*
 & / & / &   /
 & \textbf{20M}  & 115K &  \textbf{8}*
 & \textbf{20M}  & 116K  & 22*
 & / & / &   /
\\  \hline
 

\end{tabular}

 \caption{\footnotesize Summary of datasets and comparison to baselines on decomposition quality, size and running time in seconds. Decomposition results shown are for 200 timestamps in Twitch-M and Twitch-S . 
 Column $m_i$ show the size of the $i$-th tensor mode, while Prior specifies the type of side information available which in turn informs the choice of dictionary for \ourmeth and TGSD. All datasets have a graph prior for $m_1$. We explicitly denote the dictionaries used each dataset in  Fig.\ref{fig:all_decomp}. *Method time was recorded using sparse tensor representation.    }

\label{table:datasets}
\end{table*}

\subsection{Scaling to large and sparse tensors}
\label{sec:sparse_impute}

Algorithm~\ref{alg:opt} can be applied to general tensors regardless of their density (number of non-zero values) and the density of missing values for imputation. Many real-world tensors are sparse, and thus, it is desirable that \ourmeth takes advantage of such inherent sparsity to scale to large inputs without being memory-bound. For the case of no missing values, we can simply work with sparse tensor implementations without changing Alg.~\ref{alg:opt}. Specifically, for all sparse operations we can utilize the sparse tensor format in the tensor toolbox~\cite{tensortoolbox}.

When missing values are imputed, and their number is of the same order as the existing values, i.e., $nnz(\Omega)=O(nnz(\mathcal{X}))$, we need to ensure that all operations involving the full reconstructed tensor $\mathcal{D}$ retain sparsity. These operations include Step $13$ and $16$ because of the unfolding $D_i$ and Step 25 in Alg.~\ref{alg:opt}. These steps can be performed using a  matrix representation and the overall memory and running time complexity in these two steps will depend on the level of sparsity.


The update of the full tensor in Step 25, however, requires that we materialize in memory a large and dense tensor (as a dense product of factors $[[S \boxdot \Phi_1Y_1,\Phi_2Y_2,\Phi_3Y_3]]$). To retain sparsity in the update of $\mathcal{D}$ we need an imputation scheme which will only "fill in" unknown entries while leaving all sparse known entries as is. To this end we can view the full tensor $\mathcal{D}$ as a sum of two sparse tensors---the input $\mathcal{X}$ and the values to be imputed $\mathcal{M}$ as follows:  
\begin{equation}
\begin{aligned}
 \mathcal{D} =  \Omega \boxdot \mathcal{X}+( 1- \Omega) \boxdot \mathcal{M}.
\end{aligned} \label{eq:simple_impute} 
\end{equation}

This imputation scheme updates (sparse) unknown entries based on the learned reconstruction in the right hand side while ensuring that known entries remain exactly as they are in the left hand side. Updates of this form are theoretically well-justified in the case of normality (zero mean and identical variance) and independent 
residuals~\cite{parafac_base_impute}. Given these assumptions, this update can be viewed as an Expectation Maximisation (EM) approach~\cite{EM_impute} with linear convergence rate proportional to the number of missing entries~\cite{little2002statistical}. It also exploits sparsity, i.e., 
performing imputation based on Eq.~\ref{eq:simple_impute} increase the number of number of non-zero elements in $\mathcal{D}$ exactly by the number of missing entries. To endow \ourmeth with such a sparse imputation update we can simply we replace $\mathcal{M}$ with \ourmeth's reconstruction, namely:   
 
\begin{equation}
\begin{aligned}
 \mathcal{D} =    \Omega  \boxdot \mathcal{X}+ 
(1-  \Omega) \boxdot [[\Phi_1Y_1,\Phi_2Y_2,\Phi_3Y_3]].
\end{aligned} \label{eq:sparse_impute}
\end{equation}
A sparse imputation version of \ourmeth performs this update from Eq.~\ref{eq:sparse_impute} in Step 25 instead of the dense update listed in Alg.~\ref{alg:opt}. We 
follow a sparse retrieval scheme for $[[\Phi_1Y_1,\Phi_2Y_2,\Phi_3Y_3]]$, i.e., we only perform calculations for the relevant (missing) values.

\subsection{Dictionaries for MDTD}
\label{sec:dicts}
We utilize commonly adopted dictionaries for graph (GFT \cite{shuman2013emerging}) and temporal (Ramanujan \cite{tennetiTSP2015} and Spline \cite{Goepp2018Spline}) modes in our experimental evaluation. A concise summary of these bases can be found in~\cite{TGSD} . 
These dictionaries are simple and fast to construct. Specifically, the Ramanujan and Spline dictionaries take less than $1$ second to construct for all datasets and only require simple hyperparameters as input. The GFT takes slightly longer with a maximum creation time of $22$ seconds and requires the input of an network. We assume that this network information is readily available as is the case with many real datasets. We add the time cost of dictionary creation to the total running time of \ourmeth in all tables.  We choose these dictionaries to demonstrate that \ourmeth can achieve state-of-the-art performance without custom engineering or domain expertise. However, it is likely that experimenting with more potential dictionary options (e.g., Wavelet, DFT, or custom data-driven dictionaries) could result in increased performance. 
Different variations of our method are denoted by \ourmeth followed by the dictionary abbreviations. 
For example, \ourmeth with a Spline dictionary on the first mode, GFT on the second, and no dictionary for the third would be denoted as \ourmeth SG. We denote variations of the matrix dictionary decomposition baselines TGSD~\cite{TGSD} similarly.

%

\section{Experimental Evaluation}
\noindent We compare \ourmeth to baselines on (i) model quality and (ii) size, (iii) rank estimation and (iv) missing value imputation.  We preform and task and data specific grid search for all methods when appropriate.  To facilitate reproducibility we include a document  detailing the parameters selected, how they were set for each method, and additional material such as dictionary construction formulas with our code at \url{https://www.cs.albany.edu/~petko/lab/code.html}.


%

\subsection{Experimental Setup} 
\noindent \textbf{Datasets.} We employ synthetic data and three real world datasets for evaluation, including a spatial dataset (\emph{Crime}), social interactions from \emph{Reality Mining (RM)}, and data from content exchange (\emph{Twitch}).
We provide their statistics in Tbl.~\ref{table:datasets} and describe each dataset in what follows. 


\emph{$\bullet$ Synthetic Data.}
We generate 3-way synthetic datasets according to $2$ distinct GFT dictionaries generated from two stochastic block model (SBM) graphs and a Ramanujan periodic dictionary (max period $10$ and $400$ time steps). Communities in both SBM graphs contain half of all possible internal edges and an equal number of external edges. The first (smallest eigenvalue) $50$ and $30$ Laplacian eigenvectors respectively are used as dictionaries. We generate $10$ sparsely encoded factors for each mode with $75\%$ nonzero atom loadings set to uniformly random values in $[0,1]$. We form a tensor product of dictionary-encoded factors and add Gaussian noise at SNR=$20$ to the tensor. Synthetic samples and code to generate them can be found within our implementation available at \url{https://www.cs.albany.edu/~petko/lab/code.html}


\emph{$\bullet$ Twitch}~\cite{twitch} consists of followers viewing the content of streamers. An entry represents a follower watching a stream during a given hour. We select the top $5000$, $8000$, and $8000$ most active users and the top $300$, $500$ and $3000$ most active streamers from this dataset to form three versions of increasing size from this dataset: Twitch-S, Twitch-M, and Twitch-L respectively. The follower graph is based on co-viewing of the same stream with edge weights proportional to the number of hours the users co-viewed any stream. Similarly, we create a streamer graph based on shared viewership.

\emph{$\bullet$ Reality Mining (RM)}~\cite{eagle2006reality}
tracks the interactions of $94$ users at MIT where an entry represents the number of messages exchanged between a pair within a $12$ hour time-span.
We create a weighted graph based on the total number of messages exchanged and employ its GFT as a dictionary for the first two modes.

\emph{$\bullet$ Crime}~\cite{Crime} tracks the number of crimes that occurred in Chicago over $17$ years starting in $2001$. The first mode corresponds to $77$ community areas of Chicago.
Each entry in the tensor represents the number of crimes that took place in a particular community during a one hour period hour on a particular day (day slices are stacked to form the tensor).
We utilized a map of Chicago to create an associated network by connecting neighboring communities. 

\begin{figure*} [!t]
    \centering
    \subfigure [Synthetic]
    {
        \includegraphics[width=0.24\textwidth]{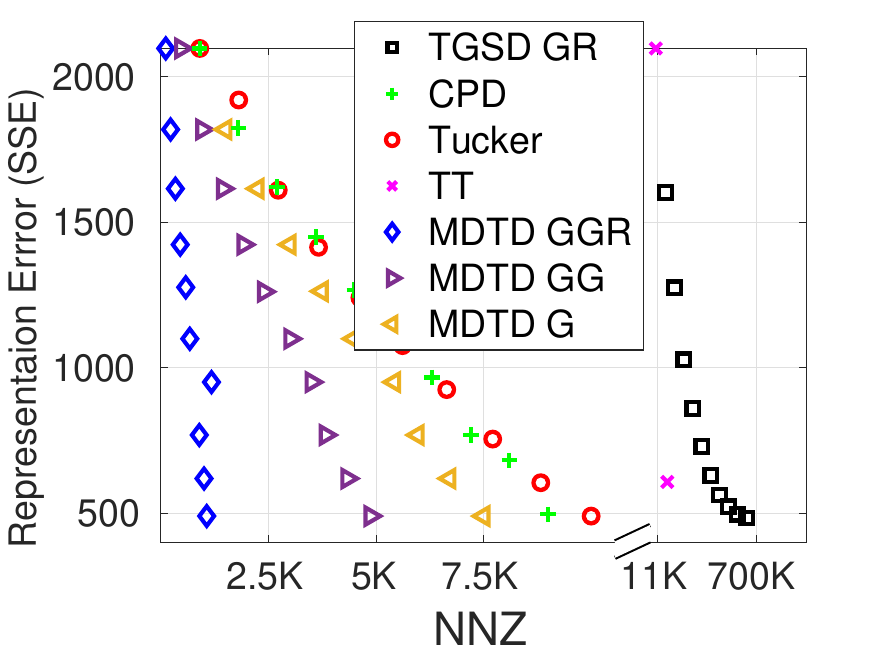}
        \label{fig:syn_decomp}
    }\hspace{-0.1in}
    \subfigure [RM]
      {
        \includegraphics[width=0.24\textwidth]{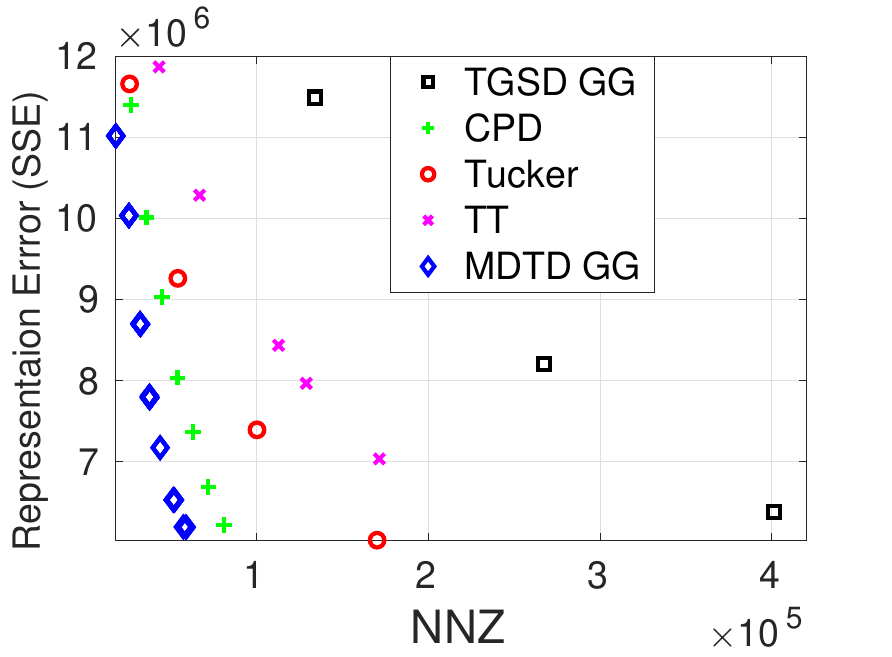}
        \label{fig:MIT_decomp}
    }\hspace{-0.1in}
     \subfigure [Twitch-S]
      {
        \includegraphics[width=0.24\textwidth]{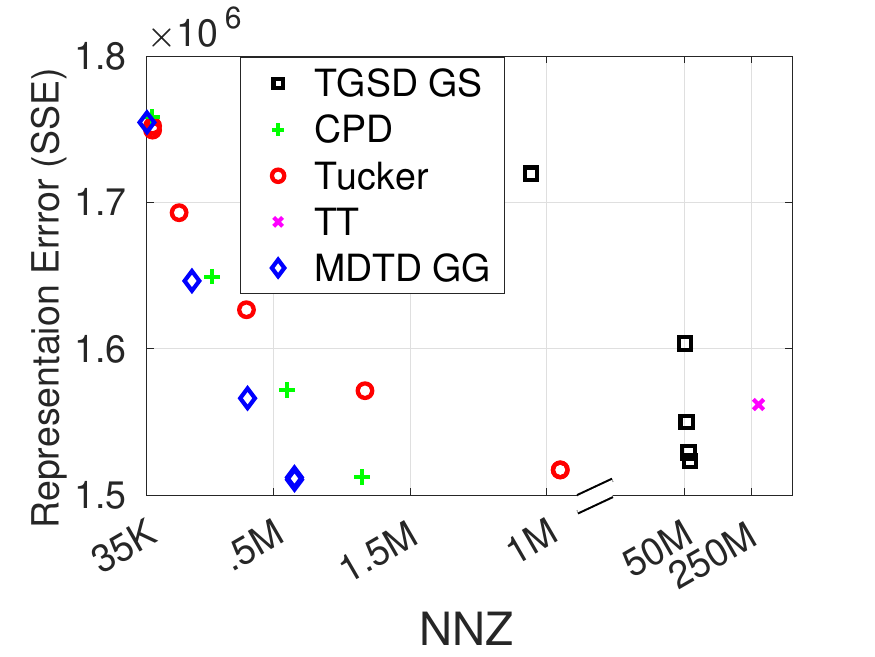}
        \label{fig:twitch_decomp}
    }\hspace{-0.1in}
   \subfigure [Synthetic Rank Estimation]
      {
     \includegraphics[width=0.24\textwidth]{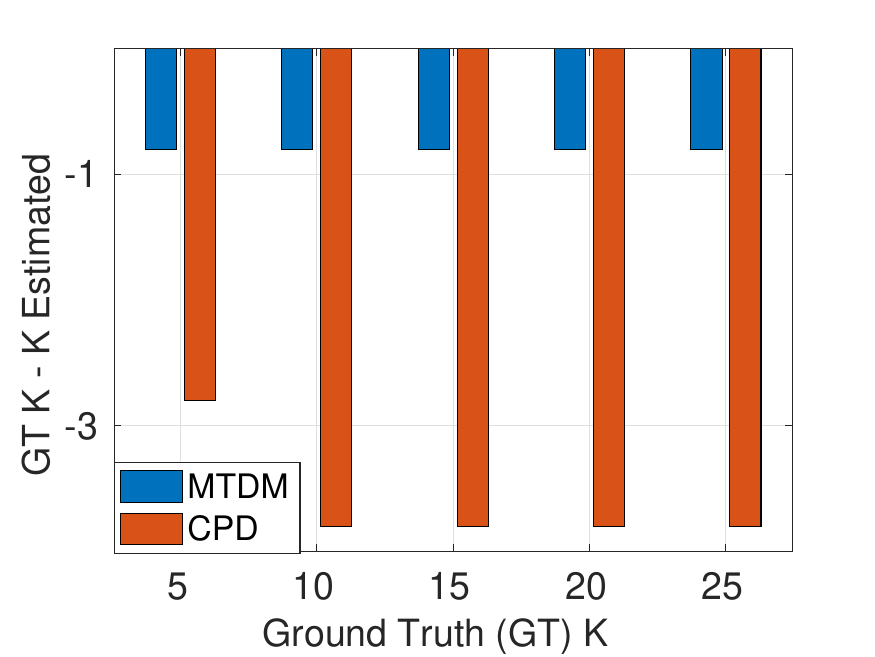}
     \label{fig:rank_estimation}
    }\hspace{-0.1in}
   \caption{\footnotesize Comparison of the quality and model size of \ourmeth and baselines CPD and Tucker on the synthetic~\subref{fig:syn_decomp}, RM~\subref{fig:MIT_decomp} and Twitch-S~\subref{fig:twitch_decomp} datasets. In synthetic, we experiment with versions of our models with increasing number of dictionaries, while for real datasets we report the best models (\ourmeth GG). Different models are obtained by varying the model rank of the competing techniques and the sparsity parameters for \ourmeth. Only Pareto-optimal models are shown in each of these figures. In Figure.\subref{fig:rank_estimation} we show the ground truth rank minus the predicted rank for various generations of the first synthetic dataset.}
       \label{fig:all_decomp}
\end{figure*}

\noindent \textbf{Decomposition baselines.}
We compare \ourmeth to CPD~\cite{PARAFAC_CPD} and Tucker decomposition~\cite{tucker1966some}, both implemented in Matlab's tensor toolbox~\cite{tensortoolbox}. We also compare to tensor train decomposition (TT)~\cite{Tensor-train}, utilizing the authors implementation. These approaches represent the state-of-the-art for low-rank tensor representation. We also compare to TGSD~\cite{TGSD}, a dictionary-based decomposition method for matrices by independetly applying it to graph-time or graph-graph tensor slices. 

\noindent\textbf{Missing value imputation baselines.}
We compare the quality of \ourmeth for missing value imputation to that of CP-WOPT~\cite{CP-WOPT} which employs CPD factorization by fitting only known values. We also compare to two Bayesian factorization approaches designed for imputation of missing values in road traffic datasets: BATF~\cite{BATF} and BCGP~\cite{BGCP}. These methods also employ a CPD-like decomposition, but regularize the factor matrix to align to Bayesian priors.
We also compare to TRLRF~\cite{TRLRF} which learns a low-rank latent space to fill in missing values; CoSTCo~\cite{costco} which utilizes a convolutional neural network to learn nonlinear dependencies among factors to impute missing values; and SOFIA~\cite{sofia}, an outlier-, seasonality-, and trend-aware tensor factorization technique for missing value imputation in temporal tensors. Finally, we also compare to TGSD~\cite{TGSD} which can impute missing matrix values and thus apply it to one tensor slice at a time.


\noindent \textbf{Baselines for tensor rank estimation.}
We utilize CPD with CONCORDIA~\cite{papalexakis2016automatic} as a baseline for tensor rank estimation.

\noindent\textbf{Metrics:} We measure quality of representation as the sum of squared error (SSE) and the quality of missing value imputation in terms of mean squared error (MSE). We quantify a model's size by the number of its non-zero (NNZ) coefficients. 
We also measure running times for each method in seconds.
COSTCO as a deep learning model was run on a Tesla V100 PCIe GPU with 16GB of RAM. All other baselines were run on a Intel(R) Xeon(R) Gold 6138 CPU @ 2.00GHz with 251G of RAM.




\subsection{Succinct decomposition}
We compare the accuracy of representation versus the size of the models when decomposing a tensor. 
We vary the decomposition rank for all methods but TT as well as the level of sparsity enforced in \ourmeth and TGSD (through the $\lambda_i$ parameters). Since Tensor-Train (TT) is capable of tuning its optimal rank for a given error level we vary the error level to obtain decompositions of varying sparsity and quality. We report the Pareto-optimal models for all methods in terms of reconstruction error (SSE) versus model size measured as NNZ. We do not count the fixed dictionary entries towards the NNZ. These dictionaries are results of preset analytical functions and can be generated efficiently on demand as discussed in Sec.~\ref{sec:dicts}. 
Tbl.~\ref{table:datasets} (right-most columns) summarizes the SSE and NNZ for one specific setting on all datasets. We select this setting by fixing a SSE level for MTDM and reporting the closest SSE regime of baselines. This allows us to compare methods in terms of model size (NNZ) for approximately similar SSE. MDTD produces the most succinct representations and its running time is comparable to the fast baselines CPD and Tucker. TT decomposes a tensor into a series of smaller tensors, leading to a typically large number of representation parameters.
While TGSD also employs dictionaries, its model sizes and running time are both larger as it is a matrix (non-tensor) baseline and cannot take advantage of 3-way dependencies in the data. Both TGSD and TT were not able to scale to Twitch-L. TT employs a dense tensor implementation and its memory needs exceed that on our experimental environment (128GB of vitrual memory) when processing the 12 billion entries in Twitch-L. While it may be possible to re-implement TT to work with sparse tensors, this is not a trivial task and is beyond the scope of the current work. TGSD on the other hand needs to preform a decomposition on a large number of slices and cannot complete in less than 24 hours.

\begin{table*}[h]
\setlength\tabcolsep{2.1pt} 
\footnotesize

 \begin{tabular}{|c|c|c|c|c|c|c|c|c| c|c|c|c|c|c|c|c|c|} 
 \hline
&  & \multicolumn{2}{c|}{\ourmeth}  & \multicolumn{2}{c|}{TGSD} &\multicolumn{2}{c|}{CP-WOPT } & \multicolumn{2}{c|}{BCGP}
& \multicolumn{2}{c|}{BATF} & \multicolumn{2}{c|}{TRLRF}  & \multicolumn{2}{c|}{CoSTCo}  & \multicolumn{2}{c|}{SOFIA} \\
\cline{2-18}  
 & \% & MSE & time   & MSE & time & MSE & time & MSE& time &  MSE & time &  MSE & time &  MSE & time   &  MSE & time    \\
\cline{2-14}  
\hline

\parbox[t]{2mm}{\multirow{5}{*}{\rotatebox[origin=c]{90}{RM}}} & 15 & \textbf{4.1} & \textbf{39}  &  \textbf{4.1} & 1K  & 17K & 1K  & 8.7 & 42K  & 250 & 3K  &  5.4 & 1K & 5.4 & 6K  &   4.7 & 1K  \\ \cline{2-18}

 & 30 & 3.7 & \textbf{34}  & \textbf{3.6} & 1K  & 21K & 1K  & 10.8 & 46K  & 50.4 & 4K  &  5.5 & 1K & 4.8 & 6K & 5.3 & 1K  \\   \cline{2-18}

 & 45 & \textbf{4.3} & \textbf{13}    &\textbf{4.3} & 3K  & 20K & 1K  & 9.2 & 55K  & 50.9 & 4K  &  7.0 & 1K  & 5.1 & 5K & 5.2 & 1K \\   \cline{2-18}

& 60 & \textbf{4.0} & \textbf{77}   & \textbf{4.0} & 4K  & 13K & 1K  & 9.4 & 55K  & 114.6 & 4K  &  7.0 & 1K & 5.1 & 3K & 5.3 & 1K   \\   \cline{2-18}

 & 75 & \textbf{4.1} & \textbf{115}   & 4.3 & 2K  & 47K & 1K  & 10.4 & 58K  & 1K & 6K  &  7.3 & 1K & 5.0 & 2K  & 5.2 & 1K \\   \cline{2-18}        \hline

\parbox[t]{2mm}{\multirow{5}{*}{\rotatebox[origin=c]{90}{Crime}}}  & 15 & .43 & \textbf{.8}  & .47 & 166 & .42 & 1K  & .51 & 7K  & \textbf{.41} & 775  &  .83 & 1K & .43 & 55K  & .50 & 36 \\  \cline{2-18}

 & 30 & .43 & \textbf{.5}  & .47 & 186 & .43 & 394  & .53 & 5K  & \textbf{.42} & 670  &  .82 & 1K& .43 & 48K & .47 & 42 \\ \cline{2-18}

 & 45 & .43 & \textbf{.7}  & .47 & 247 & .49 & 251 & .55 & 5K  & \textbf{.42} & 622  &  .86 & 1K & .43 & 33K & .45 & 39 \\  \cline{2-18}

 & 60 & .43 & \textbf{.7}  & .47 & 276 & 2.6 & 276  & .58 & 5K  & \textbf{.42} & 616  &  .97 & 1K & .43 & 24K &  .44  & 37  \\  \cline{2-18}

 & 75 & \textbf{.43 }& \textbf{.7}  & .47 & 301 & 211 & 284 & .64 & 6K  & \textbf{.43} & 668   & 1.18 & 1K &\textbf{.43 }& 15K & .45 & 33 \\  \hline

 \parbox[t]{2mm}{\multirow{5}{*}{\rotatebox[origin=c]{90}{Twitch-S}}} & 15 & $\boldsymbol{.006}$ & \textbf{1K}  & .007 & 22K & 3 & 7K  & $.009$ & 27K  &  / &  /  &  / &  /  &  / &  / & \textbf{.006} & 20K   \\  \cline{2-18}

 & 30 & $\boldsymbol{.006}$ & \textbf{1K}   & .007 & 28K & 12 & 11K  & $.008$ & 26K  &  / &  /  &  / &  /   &  / &  / & \textbf{.006} & 17K \\  \cline{2-18}

& 45 & $\boldsymbol{.006}$ & \textbf{3K}  & .008 & 40K & 40 & 15K  & $.008$ & 26K  &  / &  /  &  / &  /   &  / &  / & \textbf{.006} & 16K \\  \cline{2-18}

& 60 & $\boldsymbol{.006}$ & \textbf{9K}  & .009 & 74K & 32 & 14K  & $.009$ & 26K &  / &  /  &  / &  /  &  / &  / & \textbf{.006} & 14K  \\  \cline{2-18}

 & 75 & $\boldsymbol{.006}$ & 18K  &  / & / & 40 & \textbf{14K} & $.012$ & 26K  &  / &  /  &  / &  /   &  / &  / & \textbf{.006} & 17K \\ \cline{2-18}    \hline 
 
\end{tabular} \label{Table:missing_value}

\caption*{ \footnotesize  \textbf{TABLE II:}
  Comparison of the quality (MSE) and running time (seconds) for dense missing value imputation between \ourmeth and baselines on the real-world datasets.  \ourmeth utilizes GGS for RM and Twitch and GSS for Crime. For TGSD we report results employing the best performing GFT+spline (GS) dictionary combination across datasets. Settings in which baselines did not complete within $24$ hours are marked by the symbol "/".    }
\label{Table:missing_value_Dense}

\end{table*}

We present the full spectrum of regimes for competing techniques in Fig.~\ref{fig:all_decomp}. For our Synthetic graph-graph-time dataset we include versions of MDTD that utilize increasing set of dictionaries to serve as an ablation study evaluating the advantage of multi-dictionary decomposition~\ref{fig:syn_decomp}. Specifically, \ourmeth GGR employs all three dictionaries, \ourmeth GG employs only the graph dictionaries, while \ourmeth G employs a graph dictionary only for the first mode. Recall that factors in non-dictionary modes of \ourmeth are learned similar to regular CPD factors. The joint benefit of using multiple dictionaries for encoding is evident from this comparison. The reduction in model size is super-linear with the number of dictionaries employed at the same level of SSE.
For almost perfect fit (SSE $\approx 0$), the single dictionary version \ourmeth G requires $75\%$ of Tucker's (CPD's) coefficients, the two-dictionary version \ourmeth GG requires less than $50\%$ of those coefficients, while the 3-dictionary version \ourmeth GGR requires only $10\%$ of the coefficients. This super-linear improvement is due to the interaction of the dictionaries in the multi-way data and is also observed for sparser models of higher SSE. TGSD GR is employed on graph-time slices and is unable to utilize dependencies among all three modes leading to a huge gap in model size compared to alternatives (Note that we have interrupted the horizontal NNZ axis to enable a legible visualization including TGSD and TT).

We also compare the representation quality and model size of \ourmeth to that of CPD, Tucker, TT and TGSD on the RM and Twitch-S datasets in Figs.~\ref{fig:MIT_decomp},\ref{fig:twitch_decomp}. In both experiments we employ GFT dictionaries for two modes for \ourmeth: in the MIT they are based on a user-user social graph and for Twitch we employ weighted streamer-streamer and viewer-viewer graphs based on shared viewers and co-viewed streams respectively. Adding a temporal dictionary for the third mode in these datasets did not enable improvements on this task, indicating that the temporal behavior does not allow a significantly sparser encoding via the (Spline and Ramanujan) dictionaries we considered. It is important to note, however, that for the application of missing value imputation (Sec.~\ref{sec:missing}), the Twitch dataset benefits from a spline dictionary. The TGSD baselines is the worst among competitors since it is the only non-tensor method. Among the four baselines, CPD enables the most succinct fits. \ourmeth dominates all baselines at all levels of SSE and enables up to $5$-fold reduction of the model size compared to TGSD on the RM and Twitch datasets.


\subsection{Tensor rank estimation} 

An important parameter for all competing techniques is the selection of optimal decomposition rank. In contrast to the matrix case, determining the rank of tensors is an NP-hard problem~\cite{tensor_tutorial}. However, there exist heuristics in the literature which can be utilized to estimate the tensor rank efficiently with a popular representative: the Core Consistency Diagnostic (CCD)~\cite{papalexakis2015fast}. The central idea in CCD is to incorporate an additional tensor $(G)$ in the fit of a CPD decomposition: $\underset{G}{min} ||vec(\mathcal{X})-(A \otimes B \otimes C)vec(G) ||_F^2 $  and perform decomposition at different ranks. The rank that produces a $G$ which is closets to a super diagonal (minimum off diagonal energy) is predicted as the rank of the tensor. The intuition is that a model with proper rank will not benefit from mixing of learned factors as they individually ``cover'' the main data patterns.

We compare how well CPD and \ourmeth determine the rank of our synthetic dataset by following the CCD procedure while utilizing the implementation of~\cite{papalexakis2015fast}. We give each model a range of possible ranks: starting from $5$ less and up to $5$ more than the true rank. We then run the experiment for $5$ independently sampled version of the Synthetic dataset and report the average deviation of the estimated ranks from the ground truth. We repeat this experiment for input of increasing ranks in the range $5-25$ with a step size of $5$ and report all results in Fig.~\subref{fig:rank_estimation}. \ourmeth consistently outperforms CPD, by only slightly underestimating the ground truth rank on average. Intuitively, \ourmeth  succinctly represents ``complex'' factors that align well with dictionary atoms allowing for less cross-factor mixing at (or in the vicinity) of the ground truth rank. This makes our learned representations more distinct and representative of the underlying data generation. This in turn allows the CCD measure to identify the true rank of the data.

\subsection{Missing values imputation}
\label{sec:missing} 
We next evaluate the utility of \ourmeth for predicting missing values and compare it against baselines specifically designed for this task. We consider two scenarios: (i) \emph{dense imputation} in which a large number of up to $75\%$ of the possible tensor elements have to be imputed, and (ii) a \emph{sparse imputation} scenario in which the goal is to impute values of the same order as those present in large and sparse tensors.

\noindent {\bf Dense imputation.}
For this experiment we remove a set percentage (from $15\%$ to $75\%$) of values at random from a given tensor and then compare the accuracy of competing imputations on these held-out values measured in terms of mean squared error (MSE) and running time measured in seconds. 
To tune all methods we perform a grid search over their hyper-parameters and select the configurations which produced the smallest MSE on a validation set for all datasets with the exception of Twitch-S. We found that this dataset is too large for some competing methods to grid search their hyper-parameters extensively. To ensure a fair comparison despite this, we set the rank of all models to $50$, set \ourmeth's and TGSD's $\lambda_i=.0001$ for all $i$, and use the default parameters for other competitors. We do not report the performance on Twitch-S for methods which were not able to complete one run within $24$ hours (indicated by a ``/'' in Tbl.II).

All results from this experiment are presented in Tbl.II. Our method is consistently the best or close to the best method in terms of MSE and almost always much faster than alternatives. In RM \ourmeth is tied for the overall best performance with TGSD in terms of MSE, however, it is an order of magnitude faster. On the largest dataset in this experiment Twitch-S, \ourmeth's performance is the best in terms of both MSE and running time with the exception of when $75\%$ of the values are missing. In that regime CP-WOPT is $22\%$ faster, however, its MSE is more than 3 orders of magnitude worse. In Crime, \ourmeth is a very close second to BATF in terms MSE, but has up to $10$ orders of magnitude speed-up against the latter. Notably, for this task of missing value imputation (unlike the decomposition task) dictionary encoding for all modes resulted in optimal \ourmeth models. In particular, we employed a spline dictionary for the temporal mode in, RM, Twitch-S, and the Crime datasets, effectively enforcing smoothness in time to help impute missing values in addition to smoothness on the respective graphs associated with non-temporal modes.

\begin{table}[t!]
\setlength\tabcolsep{2.1pt} 
\footnotesize
\begin{tabular}{|c|c|c|c|c|c|c|c|c|c|c|c|} 

 \hline
& & \multicolumn{2}{c|}{\ourmeth}  & \multicolumn{2}{c|}{TGSD} &\multicolumn{2}{c|}{CP-WOPT } & \multicolumn{2}{c|}{BCGP}
& \multicolumn{2}{c|}{SOFIA} \\
\cline{2-12}    
 
 & \#t & MSE & time    & MSE  & time & MSE & time & MSE  & time & MSE & time    \\ \hline   

 \parbox[t]{2mm}{\multirow{5}{*}{\rotatebox[origin=c]{90}{Twitch-S}}}
 
 & 50  & \textbf{.008} & \textbf{499}& .019 & 2K & 34 & 9K & .075 & 7K & .009 & 49K \\   \cline{2-12}

 & 75 
& \textbf{.007} & \textbf{583}& .010 & 22K & .128 & 7K & .081 & 11K & \textbf{.007} & 55K\\   \cline{2-12}

 & 100 & \textbf{.006} & \textbf{614}& .007 & 23K & .762 & 4K & .040 & 15K & .007 & 47K \\   \cline{2-12}

 & 200& \textbf{.006} & \textbf{670}& .007 & 32K & / & / & / & / & \textbf{.006} & 43K \\   \cline{2-12}  

 & 500
& \textbf{.004} & \textbf{212}& .005 & 38K & / & / & / & / & / & /  \\    \hline

  \parbox[t]{2mm}{\multirow{5}{*}{\rotatebox[origin=c]{90}{Twitch-M}}}
 
 & 50&  \textbf{.004} &  \textbf{3K} & .008 & 93K & 16 & 6K & .025 & \textbf{ 3K} &  \textbf{.004} & 120K \\   \cline{2-12}

 & 75 &  \textbf{.005} &  \textbf{347} & .009 & 45K & .009 & 12K & .008 & 33K & .006 & 88K  \\   \cline{2-12}   

 & 100 &  \textbf{.005} &  \textbf{3K} & .006 & 76K & \textbf{.005} & 15K  & / & / & \textbf{.005 }& 113K \\   \cline{2-12}  

 & 200 &  \textbf{.006} &  \textbf{486}&  .007 & 89K  & / & / & / & / & / & /    \\   \cline{2-12}  

 & 500 &  \textbf{.010} &  \textbf{1K}& / & / & / & / & / & / & / & / \\  \hline   

  \parbox[t]{2mm}{\multirow{5}{*}{\rotatebox[origin=c]{90}{Twitch-L}}}
 
 & 50   &  \textbf{.001} &  \textbf{321}& / & / & / & / & / & / & / & /  \\   \cline{2-12}

 & 75 &  \textbf{.002} &  \textbf{3K}& / & / & / & / & / & / & / & / \\   \cline{2-12}

 & 100 &  \textbf{.001} &  \textbf{3K}& / & / & / & / & / & / & / & / \\   \cline{2-12}

 & 200 &  \textbf{.001} &  \textbf{4K}& / & / & / & / & / & / & / & / \\   \cline{2-12}  

 & 500 &  \textbf{.001} &  \textbf{1K}& / & / & / & / & / & / & / & / \\  \hline
\end{tabular} 
\label{Table:sparse_missing_value}
  \caption*{  \footnotesize \textbf{TABLE III:}
  Comparison of the quality (MSE) and running time (seconds) for sparse missing value imputation between \ourmeth and baselines on the real-world datasets.  \ourmeth utilizes GGS. For TGSD we report results employing the best performing GFT+spline (GS) dictionary combination across datasets.  Settings in which baselines did not complete within $24$ hours \emph{or} exceeded 32GB of memory are marked by the symbol "/".   } 
 
\end{table}


\begin{figure*}[t]
    \centering
    \subfigure [Syn run time vs \#nodes]
      {
        \includegraphics[width=0.22\textwidth]{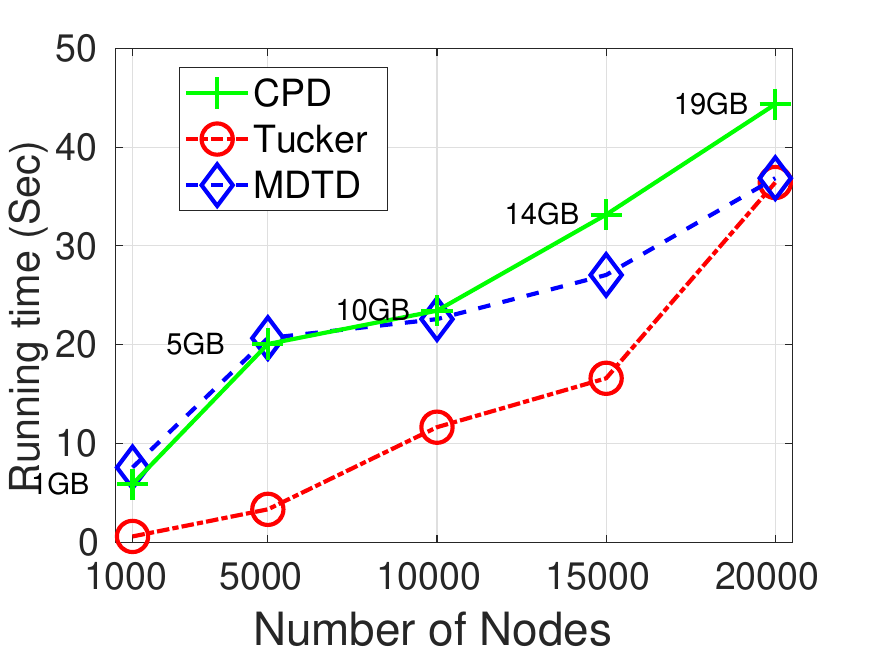}
        \label{fig:run_time_nodes}
    }\hspace{-0.1in}
             \subfigure [Syn run time vs \#timestamps]
      {
        \includegraphics[width=0.22\textwidth]{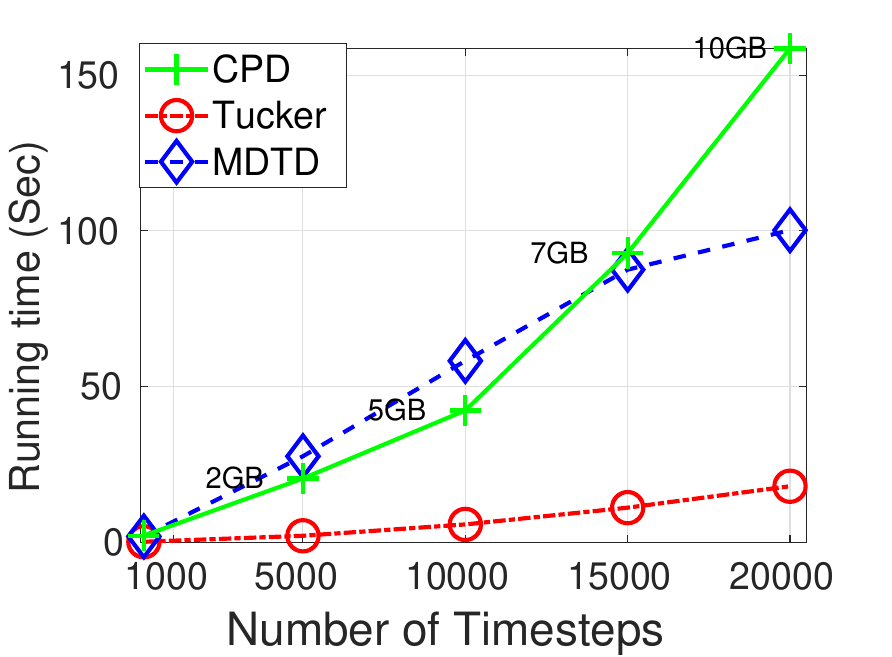}
        \label{fig:run_time_time}
    }\hspace{-0.12in}
        \subfigure [Convergence]
    {
        \includegraphics[width=.22\textwidth]{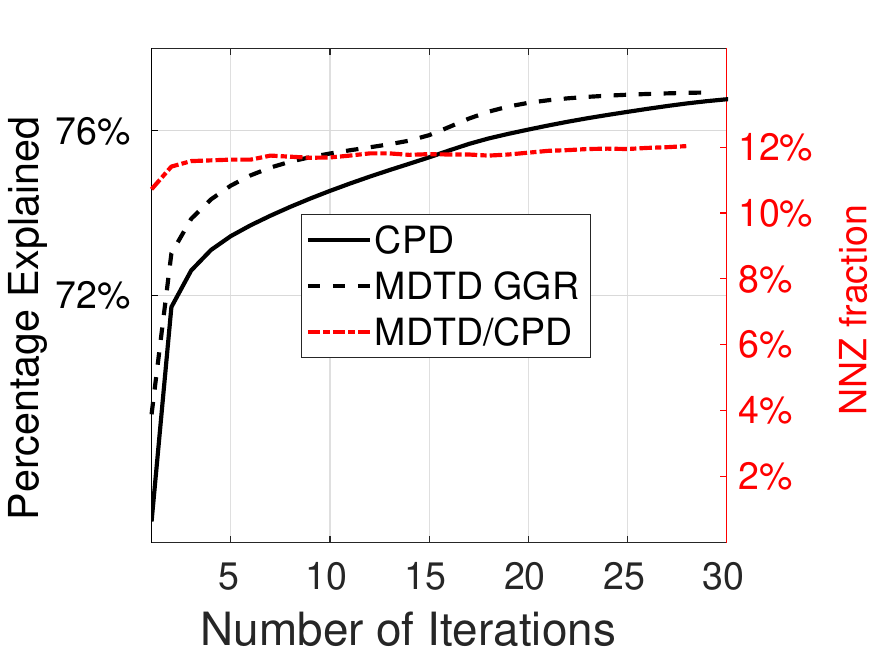}
        \label{fig:convergence_test}
    }\hspace{-0.2in}
     \subfigure [Top Twitch streamer groups]{\includegraphics[width=0.25\textwidth,clip,trim={4cm 3cm 4cm 0cm}]{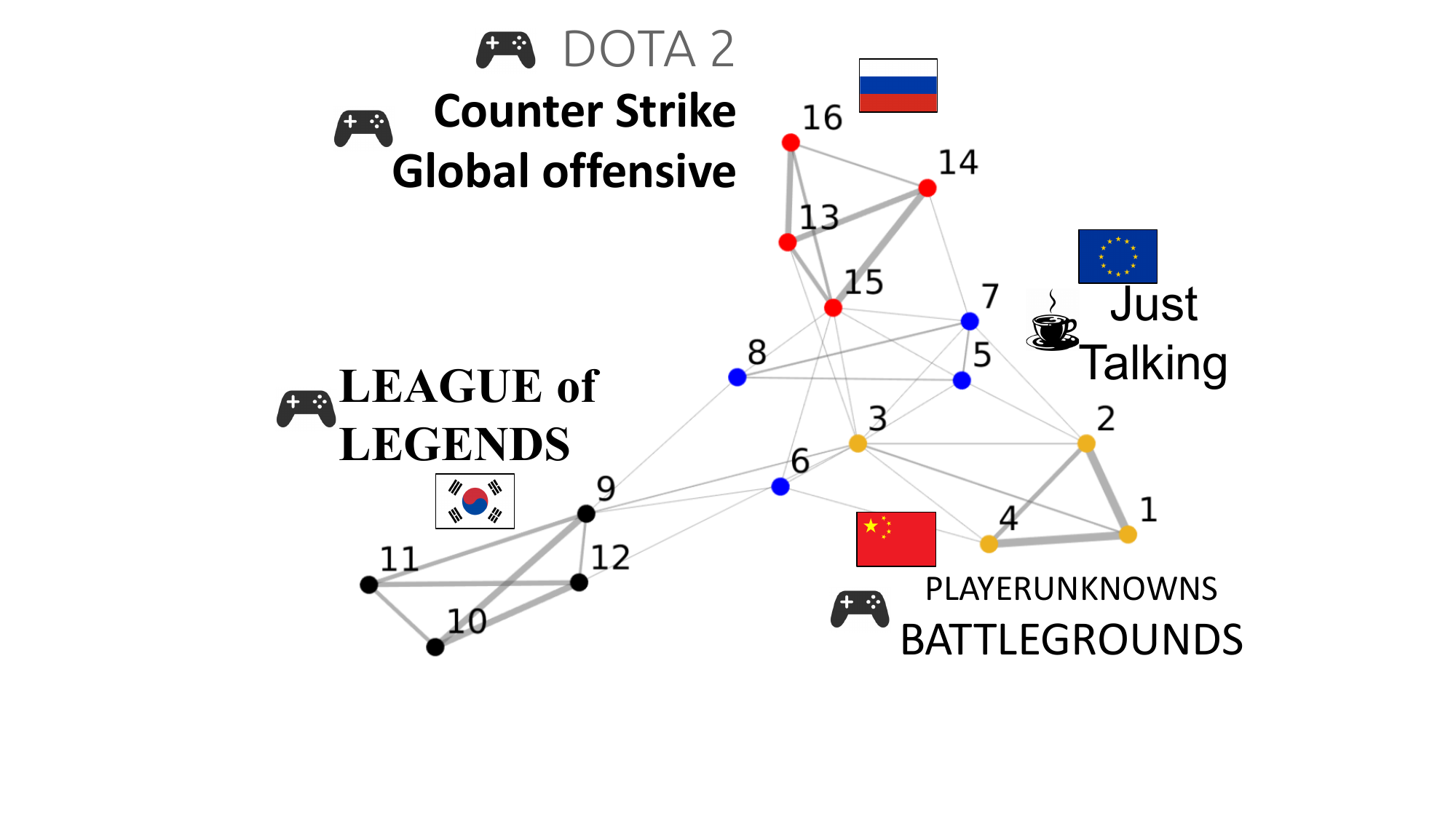}
        \label{fig:case_study}
    }

    \caption{\footnotesize 
 Scalability comparison of \ourmeth, CPD, and Tucker for increasing number of nodes \subref{fig:run_time_nodes} and timesteps \subref{fig:run_time_time} in a synthetic dataset. \subref{fig:convergence_test}: Comparison of the representation quality of \ourmeth and CPD as a function of the number of iterations of updates on a synthetic dataset. \subref{fig:case_study}:  A network among the top Twitch streamers extracted by \ourmeth. Streamers from the same tensor factors share color and are mostly co-located geographically.
  }
 
    \centering

    \label{fig:all_scalablility}
 \end{figure*}

\noindent {\bf Sparse imputation.} Next we evaluate \ourmeth's ability to to impute missing values in the large sparse tensors based on our sparse update scheme described in Sec.~\ref{sec:sparse_impute}(b). We utilize the three versions of the Twitch dataset and compare the imputation quality and scalability of competing methods that are able to scale to Twitch-S. Specifically, we vary the size of the temporal mode (number of timesteps 50 to 500), and set the number of missing values (randomly selected from all slices) equal to the number of nonzero entries in the smallest temporal length of 50 timesteps. To ensure fair comparison we limit all methods to utilize a total of 64GB of main memory.
MSE and timining results from this experiment are presented in Tbl.III . \ourmeth achieves the best performance across varying temporal lengths of the tensors. More importantly, it is able to complete imputation across all data sizes within the memory constraints. This is due to \ourmeth's ability to exploit and preserve the sparsity in both the input tensor and the set of missing values to predict. For a tensor of $12$ billion possible entries ($500$ time-steps in Twitch-L), it requires less than $17$ minutes to complete. In contrast, competing tensor methods primarily utilize dense tensor representations and quickly exhaust the available memory as the tensor grows. The baseline TGSD performs imputation in one tensor slice at a time and does not exceed the memory constraint. However, it becomes prohibitively slow on large tensors and its quality of imputation is worse than that of tensor counterparts. CP-WOPT can utilize a sparse representation, however, this variation of the method requires longer than $24$ hours to complete on the smallest dataset Twitch-S with $50$ time steps. This is because the sparse CP-WOPT is designed for imputing all possible values in the tensor (i.e., dense imputation given sparse known values).

\subsection{Scalability and Convergence}  We also compare the scalability of \ourmeth to that of CPD and Tucker models on Synthetic data. We exclude TGSD and TT from this comparison since they are both significantly slower as demonstrated in all experiments above (See Tbl.~\ref{table:datasets}). 
We record the time it takes for CPD, Tucker and \ourmeth algorithms converge under the same convergence criteria ($\epsilon=10^{-4}$). In Fig.~\ref{fig:run_time_nodes} we vary the number of nodes in the first mode while holding other modes fixed to their default sizes.
In Fig.~\ref{fig:run_time_time} we perform the same experiment only with varying the number of timestamps.
We utilize \ourmeth GGR in all settings. We also annotate the size of the input tensor in GB to illustrate the scale of the inputs considered. While Tucker is the fastest among the three competitors for small sizes, \ourmeth is a close second. As the size of the tensors grows, \ourmeth closes the gap to Tucker. For example, at 20k nodes their running times are on par. \ourmeth method is highly scalable regardless of its more complex objective and relatively less-optimized implementation (note that Tucker's and CPD's implementations well optimized library). In particular \ourmeth is able to decompose $19$ gigabyte tensors in under $1$ minute, making it applicable to large real-world datasets. 


To quantify how \ourmeth's dictionary encoding impacts the quality and speed of convergence, we track the obtained fit percentage  as a function of update iterations and compare it to that of the traditional ALS-based CPD model on our synthetic datasets. Explicitly, for fit we measure $1-\frac{||\mathcal{X}-\mathcal{R}_{con}||_f}{||\mathcal{X}||_F}$ where $\mathcal{R}_{con}$ is the reconstruction produced by each of the competing models.
We also plot the relative model size of \ourmeth v.s. CPD measured via NNZ coefficients. We terminate each algorithm when the convergence criteria of $\epsilon<10^{-4}$ is met.


We report the results in Fig.~\ref{fig:convergence_test}. The two models progress similarly in terms of the overall quality of fit per iteration but \ourmeth achieves similar fit this with less than $12\%$ of the number of coefficients employed by CPD. 
Importantly, the regularizers do not impact \ourmeth's rate of convergence comparative to CPD. 




\subsection{Case Study: Twitch's top streamer network }

We next focus on a case study to elucidate the patterns that our encoding is able to uncover. Specifically, we run \ourmeth on the Twitch-S dataset with a GFT dictionary on the first mode, a ``band-limited'' GFT on the second mode, a Ramanujan periodic dictionary on the third mode, and with $k=5,\lambda_1,\lambda_3=.0001$ and $\lambda_2=1$. We visualize the network of the $4$ streamers of highest encoding weight from $4$ of the MDTD factors learned for the streamer (second) mode of the tensor in Fig.~\ref{fig:case_study}. 
Streamer nodes from the same factors share a color and edges of very small weight (shared audience) are removed for clear presentation. We also annotate the figure by adding both the geopolitical flag associated with the origin of the streamers as well as the most popular types of streams in each group. Streamers with shared-location are grouped in \ourmeth's factors since they are also likely to share temporal patterns (stream during active hours in the same time zone) but also share audience due to cultural similarity. Because of its dictionary decomposition, \ourmeth is able to utilize such distinct temporal trends and audience network structure locality to identify groups of streamers who have similar viewers and streaming patterns. Thus, it is not surprising that these learned factors strongly align with countries. 

Simply visualizing the learned factors reveals information that may be useful to both streamers and Twitch's engineers. For example, \ourmeth has identified streamers who may be strong competitors for audience. Streamer $1$ may find it useful to know that streams $2$ and $4$ have many shared viewers who may also be interested $1$'s stream. 
Such analysis may also inform Twitch's recommendation engine. For example, it may be useful to avoid recommending cross country European streams given the small weight of edges between these streamers despite the strong association in the temporal domain. In addition, Amazon (owner of Twitch)
may be able to better provision the regional usage of its cloud servers by a better understanding of the periodic patterns of viewing encoded by \ourmeth' third-mode factors.

\section{Conclusion}
In this paper we introduced a flexible and general framework for dictionary decomposition of tensors, named \ourmeth. Our framework produced succinct low-rank representations for both synthetic and real-world tensors by jointly employing dictionaries for multiple modes in the data. We demonstrated that our proposed ADMM optimization for \ourmeth converges to a high quality solution on par with CPD and Tucker in many settings. Moreover, the resulting factors were shown to be advantageous through their utility for succinct representation, their capability of estimating the ground truth rank, and their ability to accurately model the underlying patterns in the data in the presence of missing values. Our code and sample synthetic datasets are available at \url{https://www.cs.albany.edu/~petko/lab/code.html}.
%

\balance
{
\bibliographystyle{abbrv}
\bibliography{references}
}





\end{document}


\title{Multi-Dictionary Tensor Decomposition\\
         \Large Reproducibility Materials}


\author{Anonymous}
\date{}

\maketitle




\maketitle

\input{text/reproducibilityICDM}
{
\footnotesize
\balance
\bibliographystyle{siam}
\bibliography{references}
}